\documentclass[3p]{elsarticle}

\clearpage{}\usepackage{ifthenx}

\makeatletter{}
\newboolean{classArticle}
\@ifclassloaded{article}{
  \setboolean{classArticle}{true}
}{
  \setboolean{classArticle}{false}
}

\newboolean{classElsarticle}
\@ifclassloaded{elsarticle}{
  \setboolean{classElsarticle}{true}
}{
  \setboolean{classElsarticle}{false}
}
\makeatother{}
\clearpage{}\usepackage[a-1a]{pdfx}\clearpage{}
\clearpage{}\usepackage{ifluatex}
\ifluatex
  \usepackage{polyglossia}
  \setmainlanguage{english}
\else
  \usepackage[T1]{fontenc}
  \usepackage[utf8]{inputenc}
  \usepackage[english]{babel}
\fi
 \usepackage{amssymb,amsfonts,amsmath}
\usepackage{siunitx}
 \usepackage{xparse}
\usepackage{csquotes}
\ifluatex
  \usepackage{biblatex}
  
  \DeclareFieldFormat{eprint:hal}{
    \mkbibacro{HAL}\addcolon\space
    \ifhyperref
    {\href{https://hal.archives-ouvertes.fr/#1}{\nolinkurl{#1}}}
    {\nolinkurl{#1}}}
  \DeclareFieldFormat{eprint:openreview}{
    OpenReview\addcolon\space
    \ifhyperref
    {\href{https://openreview.net/forum?id=#1}{\nolinkurl{#1}}}
    {\nolinkurl{#1}}}
  
  \providecommand{\citet}[1]{\textcite{#1}}
  \providecommand{\citep}[1]{\parencite{#1}}
\else
  \usepackage{natbib}
  \providecommand{\parencite}{\citep}
  \providecommand{\textcite}{\citet}
\fi
 \usepackage{rotating}
\usepackage{booktabs}
\usepackage{multirow}
\usepackage{xcolor}
\definecolor{LRed}{HTML}{FEE5D9}
\definecolor{MRed}{HTML}{FB6A4A}
\definecolor{DRed}{HTML}{A50F15}
\definecolor{LGreen}{HTML}{EDF8E9}
\definecolor{MGreen}{HTML}{74C476}
\definecolor{DGreen}{HTML}{006D2C}
\definecolor{LBlue}{HTML}{EFF3FF}
\definecolor{MBlue}{HTML}{6BAED6}
\definecolor{DBlue}{HTML}{08519C}
\usepackage{fontawesome}
\usepackage{subcaption}
\usepackage{algorithm}
\usepackage{algpseudocode}
\floatname{algorithm}{Algorithm}

\usepackage{hyperref}
\hypersetup{
  unicode=true,
  pdfusetitle,
  pdfborder={0 0 0},
  breaklinks=true,
  colorlinks=true
} \clearpage{}
\clearpage{}
\newlength{\myInclusionWidth}
\newlength{\myInclusionHeight}

\providecommand{\includesvg}[3][\columnwidth]{
  \setlength{\myInclusionWidth}{#1}%
  \setlength{\myInclusionHeight}{0.625\myInclusionWidth}%
  {%
    \graphicspath{{#2/}}%
    \resizebox{#1}{!}{\input{#2/#3.pdf_tex}}%
  }%
}
\providecommand{\includeplot}[3][\columnwidth]{
  \setlength{\myInclusionWidth}{#1}%
  \setlength{\myInclusionHeight}{0.625\myInclusionWidth}%
  {%
    \graphicspath{{#2/}}%
    \resizebox{#1}{!}{\input{#2/#3.tex_plot}}%
  }%
}
 
\providecommand{\ARCADES}[1]{\textsc{Arcades}#1}

\DeclareMathOperator*{\argmax}{argmax}

\providecommand{\myProjectWeb}[1]{%
  \href{https://brenona.gricad-pages.univ-grenoble-alpes.fr/arcades/}{\nolinkurl{brenona.gricad-pages.univ-grenoble-alpes.fr/arcades/}#1}
}

\providecommand{\arcadesExpeResults}[8]{
  \typeout{^^JExpe results: #1...^^J}
  \begin{minipage}{\columnwidth}
    \centering
    \begin{subtable}[c]{\textwidth}
      \centering
      \caption{Final scores}
      \label{tab:#1:scores}
      \setlength{\tabcolsep}{1.5em}
      \begin{tabular}{cccc}
        \toprule{}
        Reward per ep. & Precision & Recall & F1-score \\%
        \midrule{}
        #2 & #3\thinspace{}\si{\percent} & #4\thinspace{}\si{\percent} & #5\thinspace{}\si{\percent}\\%
        \bottomrule{}
      \end{tabular}
    \end{subtable}
    \bigskip{}
    
    \begin{subfigure}[t]{0.49\textwidth}
      \centering{}
      \includeplot{#8}{rewardPerEp}
      \caption{\emph{Average reward per episode} growth}
      \label{fig:#1:rewardPerEp}
    \end{subfigure}
    \hfill{}
    \begin{subfigure}[t]{0.49\textwidth}
      \centering
      \includeplot{#8}{f1Score}
      \caption{\emph{F1-score} growth}
      \label{fig:#1:f1score}
    \end{subfigure}
    \caption[#6]{#7}
    \label{fig:#1}
  \end{minipage}
  \typeout{^^JExpe results: #1... DONE^^J}
}

\NewDocumentCommand{\targetq}{m}{\mbox{\texttt{target\_q}}#1}
\NewDocumentCommand{\minibatchsize}{m}{\mbox{\texttt{minibatch\_size}}#1}
\NewDocumentCommand{\updatefreq}{m}{\mbox{\texttt{update\_freq}}#1}

 \newcounter{architecture}
\renewcommand{\thearchitecture}{A\arabic{architecture}}

\newcounter{scenario}
\renewcommand{\thescenario}{\arabic{scenario}}

\newenvironment{scenario}{%
  \begin{small}
    \begin{quote}%
      \refstepcounter{scenario}%
      \textbf{Scenario \thescenario}%
      \quad
      \itshape
    }{%
    \end{quote}%
  \end{small}
}

\makeatletter{}
\NewDocumentCommand{\@gittag}{}{}
\NewDocumentCommand{\@data}{}{}
\NewDocumentCommand{\@totalsteps}{}{}
\NewDocumentCommand{\@learningsteps}{}{}
\NewDocumentCommand{\@evalsteps}{}{}
\NewDocumentCommand{\gittag}{m}{\RenewDocumentCommand{\@gittag}{}{#1}}
\NewDocumentCommand{\data}{m}{\RenewDocumentCommand{\@data}{}{#1}}
\NewDocumentCommand{\totalsteps}{m}{\RenewDocumentCommand{\@totalsteps}{}{#1}}
\NewDocumentCommand{\learningsteps}{m}{\RenewDocumentCommand{\@learningsteps}{}{#1}}
\NewDocumentCommand{\evalsteps}{m}{\RenewDocumentCommand{\@evalsteps}{}{#1}}
\NewDocumentEnvironment{ArcadesConfig}{s o o}{
    \IfBooleanTF{#1}{%
      \begin{subtable}[#2]{#3}%
    }{%
      \begin{table}[#2]%
    }%
    \centering%
  }{%
    \begin{tabular}{rp{95pt}}%
      \toprule%
      Parameter & Value \\%
      \midrule%
      Git tag & \texttt{\@gittag{}}\\%
      Data & \@data{} \\%
      \midrule%
      $totalLearningSteps$ & \@totalsteps \\%
      $learningSteps$ & \@learningsteps \\%
      $evaluationSteps$ & \@evalsteps \\%
      \bottomrule%
    \end{tabular}%
    \gittag{}%
    \data{}%
    \totalsteps{}%
    \learningsteps{}%
    \evalsteps{}%
    \bigbreak{}%
    \IfBooleanTF{#1}{%
      \end{subtable}%
    }{%
      \end{table}%
    }%
}
\makeatother{} \clearpage{}

\begin{document}
  \clearpage{}\ifthenelse{\boolean{classElsarticle}}{
  \begin{frontmatter}
}{}

\title{\ARCADES{}: A deep model for adaptive decision making \linebreak in voice controlled smart-home}

\ifthenelse{\boolean{classElsarticle}}{
  \author[uga]{Alexis \textsc{Brenon}}
  \ead{alexis.brenon@univ-grenoble-alpes.fr}
  \author[uga]{François \textsc{Portet}\corref{cor1}}
  \ead{francois.portet@univ-grenoble-alpes.fr}
  \author[uga]{Michel \textsc{Vacher}}
  \ead{michel.vacher@univ-grenoble-alpes.fr}
  
  \cortext[cor1]{Corresponding author}
  
  \address[uga]{Univ. Grenoble Alpes, CNRS, LIG, F-38000 Grenoble France}
}{}
 \begin{abstract}
  In a voice controlled smart-home, a controller must respond not only to user's requests but also according to the interaction context.
  This paper describes \ARCADES{}, a system which uses deep reinforcement learning to extract context from a graphical representation of home automation system and to update continuously its behavior to the user's one.
  This system is robust to changes in the environment (sensor breakdown or addition) through its graphical representation (scale well) and the reinforcement mechanism (adapt well).
  
  The experiments on realistic data demonstrate that this method promises to reach long life context-aware control of smart-home.
\end{abstract}

\ifthenelse{\boolean{classElsarticle}}{
  \begin{keyword}
    Smart-home \sep Decision system \sep Context-aware \sep Reinforcement learning \sep Deep learning
  \end{keyword}
}{}
 
\ifthenelse{\boolean{classElsarticle}}{
  \end{frontmatter}
}{
  \ifthenelse{\boolean{classArticle}}{
    \maketitle
  }{}
}\clearpage{}
  
  \clearpage{}\section{Introduction}
\label{sec:introduction}

One of the applications in which Ubiquitous Computing \parencite{Weiser1991} generates great expectations is \emph{smart-home}.
Smart-home is an application domain which brings together home automation and \emph{ambient intelligence} to ease life of dwellers and to provide support to people in loss of autonomy.
The development of smart-homes is not only a cultural and technological evolution but is also recognized as one way to address the challenges created by an aging population in developed countries \parencite{PeetoomEtAl2014}. 
If home automation is concerned with sensing (sensors, actuators, middle-ware) and low-level automation (heating control, lighting control), \emph{Ambient Intelligence} should provide perception and reasoning capabilities into the smart-home ecosystem.

However, although the development of smart-homes is supported by a large amount of research and industrial projects, it has not reached a large public since many challenges are still to be addressed.
One of the main challenges is due to the complexity of setting up the smart-home system in case of new situations (devices, house, dwellers, after an accident, etc.). 
Furthermore, smart-homes often need to perform inferences from data in order to make decision so as to provide services that enhance users' experience and control over their environment \parencite{Augusto2013}. To make this enriched control possible, these systems perceive their environment and decide which actions to apply to the environment. This decision can be made after some specific events such as a user's request, a scheduled event (e.g., programmed task) or a detected risky situation. This perception is not only useful to trigger a decision but also to adapt the decision making to the current circumstances in which such decision must be performed. These circumstances are called \emph{the context} and systems that explicitly take the context into account, are said \emph{context-aware} \parencite{Loke2006}. However, defining explicitly what pieces of information is playing an important role in the context of a decision is a difficult and tedious task. Indeed, if a general definition of context exists in the field \parencite{Dey2001}, the large amount of existing models \parencite{Bolchini2007,Ibarra2016} demonstrates not only the importance of this notion for research but also the lack of a unified way of modelling context in intelligent environments. 

Even though a model could be defined for a specific home, successful interaction must match users' preferences. To do so, explicit user configuration is a way to handle preferences but this have been proven to reduce the acceptance of the system~\parencite{ZaidenbergEtAl2010}. More subtle ways of adapting system to users' preferences must be found. Moreover, as shown by~\textcite{ZaidenbergEtAl2010}, users ask for a way to monitor the system to better understand what it does and why. Indeed, smart-home systems must let the user keep control of the house even when she is in loss of autonomy~\parencite{PortetEtAl2013}.
All these factors make the development of an end-to-end smart-home system challenging.

In this paper, we propose a new system called \ARCADES{} (Adaptive Reinforced Context-Aware Deep Decision System) to perform adaptive context-aware decision for voice command in a smart home that does not use an explicitly defined context. Rather, the context information is directly learned from data using deep neural network \parencite{Schmidhuber2015,LeCunEtAl2015}. Indeed, one of the main advantages of Deep Learning (DL) is to learn attributes from raw data at the same time as the classification model is trained. This ability of DL has been demonstrated in numerous classification tasks reaching superhuman accuracy in the case of image classification \parencite{LeCunEtAl2015}.

Another aspect of context awareness is to be able to capture a changing environment over time. Indeed, sensors, activity places, dwellers, etc. can change and any system that targets a lifelong support must be able to take these changes into account. \autoref{sen:one} illustrates this need. In this simple scenario, it is impossible to know how to act without considering the context since the two voice commands are semantically the same. Here the activity or the location influence the choice of the lamp to light. Changes in the daily living routine (changing a lamp of place, moving the table, etc) must be accounted for in the configuration of the system. If static rule based systems are employed, any change must be impacted in the rule based system, making this impossible to modify by non-technical person such as the elderly users. To adapt continuously to a changing environment, the decision model we present in this paper uses reinforcement learning to update its decision policy.

\begin{scenario}
  \label{sen:one}
  The inhabitant is washing up dishes and utters the voice command \enquote{Turn on the light}. The lamp above the sink is lighted. Then she goes to the kitchen table and utters \enquote{Give me some light}. The lamp above the table is lighted. 
\end{scenario}

Finally, to make the smart home system easy to interact with we rely on speech interaction as well as on the sensor data capturing the interaction of the user with her own home. This make the set-up accessible since the user does not have to learn a complex jargon or to deeply understand the internal mechanisms of the system. 
Relying on machine learning paradigms, \ARCADES{} is able to build a model relevant for the given task, and then to adapt itself, in a non-intrusive manner.
Moreover, we advocate the use of a new graphical approach to represent the environment state that brings two advantages: it handles multiple heterogeneous sensor data (making it resilient to smart-home evolution) and it allows users to check what the system perceives (cf.~\parencite{ZaidenbergEtAl2010}).
Thus, the work presented in this paper brings the following contributions. 
\begin{itemize}
  \item A system which represents the decision-making task as a reinforcement learning one, to adapt a decision making to the user's behaviour and her environment in an on-line setting (i.e., data are processed only with a limited history of past and no information about the future) in which the task is to perform voice command execution.
  \item A unified representation of the smart-home heterogeneous data sources, namely image, which is readable by machine and human ; hence, the model does not depend on intermediate representation layers but only on the input data. This is a strong strength of our approach as well as an original aspect of the research presented here (we are not aware of any work using this original data representation in this field).
  \item A deep learning model able to build and extract relevant contextual features from (semi-)raw sensors input (i.e. images) as well as solving the reinforcement learning task. 
  \item An evaluation of the system on a realistic smart-home data set available to the community.
\end{itemize}

The paper is organised as follows.
The next section (\autoref{sec:related_works}) summarises the related work.
The method is detailed in \autoref{sec:methods} which describes the overall approach and the data representation.
In \autoref{sec:experiment} we present an experiment to train and evaluate \ARCADES{} on realistic data.
Complementary experiments are summed up in \autoref{sec:complementary}.
Finally, the results and some possible improvements are discussed in \autoref{sec:discussion}.
\clearpage{}
  \clearpage{}\section{Related works}
\label{sec:related_works}

Automatic adaptive decision-making in smart-homes has received far less attention than more widespread research on context or situation recognition.
However, it had recently regained interest, particularly in the domain of energy management \parencite{LiEtAl2014,BerlinkEtAl2015}.
In the literature, a decision-making system depends on two main abilities: (1) the ability to represent knowledge perceived from the environment and (2) the ability to make decision from this perception given a specific user and situation.

With respect to the first ability, research on context aware systems had generated a vast amount of knowledge representations.
Logical approaches are generally applied for representation in pervasive environments because they can easily model relations among entities and perform automatic inference.
For instance, \textcite{MileoEtAl2011} applied a logic programming approach, Answer Set Programming, in order to implement a system of prediction of risky situations.
Event calculus, a method to model situations using first order logic, has also been applied in the development of smart-homes \parencite{ChenEtAl2008,KatzourisEtAl2014}.

The main trend for knowledge representation in smart environments has become ontology, in particular those based on Description Logic (DL).
The Ontology Web Language (OWL) is the main implementation of DL, and it has been largely applied in pervasive environments \parencite{RodriguezEtAl2014a,LiaoEtAl2007}.
\textcite{LiaoEtAl2007} employed an ontology to organize temporal events and the concepts about the smart-home.
Context-aware systems have also been developed using ontologies to model the current context for mobile applications \parencite{AttardEtAl2013,YilmazEtAl2012}.
Fuzzy ontologies also received attention because of their ability to model imprecision and allow inference with vague concepts such as \enquote{hot weather}.
\textcite{RodriguezEtAl2014} presented an application of fuzzy ontologies for activity recognition in pervasive environments.
Another important contribution to make ontologies capable of dealing with uncertainty is the use of probabilistic models in ontology inference.
A notable application of such approach has been presented by \textcite{HelaouiEtAl2013}.
In \textcite{ChahuaraEtAl2016} a Markov Logic Network together with an ontology in Description Logic (DL) is employed to perform context recognition.
However, the diversity of approaches also shows the lack of consensus regarding the different concepts to be considered in a smart-home and their organization.
Hence, there is a need for alternative approaches to handle information being inferred from the sensor network of the smart-homes.

In all of these approaches, context modelling is considered from a top-down perspective, where elements of the context are identified by expertise. In this paper, we propose a different approach: a data-driven constitution of the context where all the sensor data are unified into a common representation: an image. The closest work to this approach is the recent one by \textcite{Singh2017} which transforms a pressure sensor matrix image to perform gait identification using deep neural network. However, in their work the data were already sequences of 2-dimensional values while in the case of smart home we must dealt with a set of monodimensional sensor data of variable type (Boolean, categorical, numerical) whose position is only loosely known and changing.

Regarding the decision-making process, several logic-based approaches are presented in the literature.
\textcite{MooreEtAl2011} developed a system that exploit a set of fuzzy rules in order to find the most appropriate action under a certain condition given by the context.
Similarly, \textcite{KoflerEtAl2012} and \textcite{Gomez-RomeroEtAl2012} used Description Logic to define the behavior of a context-aware system which models context elements (activities) in an ontology.
There are several works applying ECA (Event-Condition-Action) rules based systems into pervasive environments \parencite{LeongEtAl2009, YauEtAl2006}.
However, in these proposals the system is set \emph{a priori} to execute an action given a specific configuration (the condition) and consequently the system does not adapt its behavior to a changing situation.
Bayesian networks, for instance, are among the most important methods for decision-making, as exemplified by \textcite{LeeEtAl2012}.
They have modeled uncertainty on context-aware system by means of Bayesian networks but the method does not include formal elements of decision-making.
Influence Diagrams (ID, \cite{HowardEtAl1981}) is a method based on Bayesian networks that includes special variables to appropriately model the decision process.
Some research work on context aware systems \parencite{DeCarolisEtAl2004,NishiyamaEtAl2011} have relied on ID in order to model the decision process, treating the uncertainty and measuring the expected utility of possible actions.
However, the expressiveness of this probabilistic approach is limited since only propositional variables and conditional dependencies among variables can be represented in the model, besides the fact that it is less human readable than logical approaches.
In \textcite{ChahuaraEtAl2017} a Markov Logic Network (MLN) is proposed to perform automatic decision-making in smart-home.
MLN make it possible to benefit from a formal representation of the decision process using first order logic while modeling uncertainty by weighting the logic formulae.
The other advantage is that weights can be learned from data.
The approach has been tested in a realistic setting and showed the strength of the approach.
However, it does not include mechanism to adapt the decision model to new situations.
Furthermore, if part of the model can be learned from data, the logic part is acquired by human expertise which has the same drawback as the logic based knowledge representations in smart-home.

To tackle the problem of adaptation, some researches in smart-home have designed the decision problem as a reinforced decision-making problem.
The most famous of these works is \emph{The Neural Network House} \parencite{Mozer1998}.
It is a house equipped with more than 75 sensors and actuators monitoring the environmental conditions --- such as temperature, light and sound levels, motions, door and windows openings --- and controlling some devices of the house such as heaters or lights.
The devices were controlled by the ACHE system \parencite{Mozer2005}, which commanded the lights in order to reduce the overall energy consumption without impairing the inhabitant comfort.
Reinforcement Learning (RL) is the base of the ACHE system.
A state estimator builds a high-level environmental state from the sensor data.
This state contains the current activity of the user and the level of natural light at the user's location.
Then, a controller makes a decision about which light to act on, and decides on the intensity level to set.
To evaluate this decision, two kinds of cost are defined and returned to the controller for its learning.
On the one hand, the energy cost, directly indexed on the actual energy cost of electricity in the area of the smart-home.
On the other hand, the comfort cost, which is induced when the inhabitant have to modify the decision made by the system.
Although this project showed promising results, it seems to be no longer active but has given rise to a number of researches by other groups \parencite{ZaidenbergEtAl2011, KaramiEtAl2016} which showed that RL allows some flexibility that traditional machine learning paradigms do not.
Nevertheless, if RL is particularly suitable for restricted environment (with a few possible states and actions), it poorly scales to large real cases sets.

This drawback, is mainly caused by the classical implementations used for reinforcement learning tasks, based on an associative array whose size depends on the number of  variables and their domain values to represent the state of the environment.
A way to solve this, is to change the data structure, relying for example on a deep neural network \parencite{LeCunEtAl2015,Schmidhuber2015}.
A notable example of \emph{deep reinforcement learning} for decision-making is provided by \textcite{MnihEtAl2015} who proposed a new approach to train a deep neural network using reinforcement learning.
The proposed model, the Deep Q-Network (DQN), has been applied to a decision-making task regarding classic Atari2600 games.
Roughly speaking, given a pre-processed image frame from the Atari display, the model chooses the best action out of 18 (17 combinations of a joystick plus 1 no action) to maximize its reward based on the game score.
A convolutional neural network (CNN) processes the frames, while a fully connected network approximates the value of the states of the RL problem.
After few millions training iterations, the model is compared to random and human decision-making and shows that the DQN reaches or surpasses human performances in the majority of tasks.
This work proves that RL and a deep convolutional network can be combined.
Thus, RL can be used in situations approaching real-world complexity.
This inspired us to use a similar agent to make context-aware decision in a smart-home.

\typeout{^^J Landscape context table...}
\begin{sidewaystable*}[p]
  \begin{scriptsize}
  \centering
  \caption{Summary of the studies related to context representation and decision-making in smart homes}
  \label{tab:related_works:sumup}

  \providecommand{\iconcrossmark}{\textcolor{DRed}{\faicon{times}}}
  \providecommand{\iconcheckmark}{\textcolor{DGreen}{\faicon{check}}}

  \renewcommand{\arraystretch}{2.1}
  \newlength{\myAboveRuleSep}\setlength{\myAboveRuleSep}{0.33em}
  \setlength{\aboverulesep}{\arraystretch\myAboveRuleSep}
  \hbadness 10000\relax{} 
  
  \newlength{\mycategorycolwidth}\setlength{\mycategorycolwidth}{10pt}
  \newlength{\mypubcolwidth}\setlength{\mypubcolwidth}{80pt}
  \newlength{\myctxtcolwidth}\setlength{\myctxtcolwidth}{100pt}
  \newlength{\mydecisioncolwidth}\setlength{\mydecisioncolwidth}{100pt}
  \newlength{\mydatacolwidth}\setlength{\mydatacolwidth}{70pt}
  \newlength{\mymarkcolwidth}\setlength{\mymarkcolwidth}{10pt}
  \newlength{\myusecolwidth}\setlength{\myusecolwidth}{70pt}
  
  \hfill
  \begin{tabular}{
      m{\mycategorycolwidth}|
      m{\mypubcolwidth}
      m{\myctxtcolwidth}
      m{\mydecisioncolwidth}
      m{\mydatacolwidth}
      *{5}{m{\mymarkcolwidth}@{\hspace{0.5\mymarkcolwidth}}} @{\hspace{2\mymarkcolwidth}}
      m{\myusecolwidth}
    }
    \toprule 
    \multicolumn{1}{c}{} &
      Publication &
      \parbox{\myctxtcolwidth}{Inferred\\contextual information} &
      \parbox{\mydecisioncolwidth}{Decision-making\\method} &
      \parbox{\mydatacolwidth}{Data\\representation} &
      \rotatebox{90}{Voice interface} &
      \rotatebox{90}{Continuous adapt.} &
      \rotatebox{90}{On-line} &
      \rotatebox{90}{Smart-home} &
      \rotatebox{90}{Multi-room} &
      Use case\\%
    \midrule{}
    \multirow{4}{=}[-3pt]{\centering\rotatebox[origin=c]{90}{No \tiny decision-making}} &
      \parbox{\mypubcolwidth}{\citeauthor{MileoEtAl2011}\\(\cite*{MileoEtAl2011})} &
      \parbox{\myctxtcolwidth}{Location,\\identity, objects} &
      -- &
      \parbox{\mydatacolwidth}{Answer set\\ programming} &
      \iconcrossmark{} &
      \iconcrossmark{} &
      \iconcrossmark{} &
      \iconcrossmark{} &
      -- &
      Synthetic data\\
    & \parbox{\mypubcolwidth}{\citeauthor{KatzourisEtAl2014}\\ (\cite*{KatzourisEtAl2014})} &
      Activity &
      -- &
      Logical &
      \iconcrossmark{} &
      \iconcrossmark{} &
      \iconcrossmark{} &
      \iconcrossmark{} &
      -- &
      Synthetic data\\
    & \parbox{\mypubcolwidth}{\citeauthor{RodriguezEtAl2014}\\ (\cite*{RodriguezEtAl2014})} &
      Activity &
      -- &
      Fuzzy ontology &
      \iconcrossmark{} &
      \iconcrossmark{} &
      \iconcheckmark{} &
      \iconcrossmark{} &
      -- &
      Synthetic data\\
    & \parbox{\mypubcolwidth}{\citeauthor{HelaouiEtAl2013}\\ (\cite*{HelaouiEtAl2013})} &
      Activity &
      -- &
      \parbox{\mydatacolwidth}{Probabilistic\\ontology} &
      \iconcrossmark{} &
      \iconcrossmark{} &
      \iconcheckmark{} &
      \iconcrossmark{} &
      -- &
      Synthetic data\\%
    \midrule{}
    \multirow{5}{=}[-3pt]{\centering\rotatebox[origin=c]{90}{Decision-making}} &
      \parbox{\mypubcolwidth}{\citeauthor{LiaoEtAl2007}\\ (\cite*{LiaoEtAl2007})} &
      Predefined situations &
      \parbox{\mydecisioncolwidth}{Dangerosity level} &
      Ontology &
      \iconcrossmark{} &
      \iconcrossmark{} &
      \iconcrossmark{} &
      \iconcrossmark{} &
      \iconcrossmark{} &
      Synthetic data\\%
    & \parbox{\mypubcolwidth}{\citeauthor{KoflerEtAl2012}\\ (\cite*{KoflerEtAl2012})} &
      Predefined situations &
      \parbox{\mydecisioncolwidth}{Ontology\\reasoning} &
      Ontology &
      \iconcrossmark{} &
      \iconcrossmark{} &
      \iconcrossmark{} &
      \iconcrossmark{} &
      -- &
      None\\
    & \parbox{\mypubcolwidth}{\citeauthor{Gomez-RomeroEtAl2012}\\(\cite*{Gomez-RomeroEtAl2012})} &
      \parbox{\myctxtcolwidth}{Activity,\\predefined situations} &
      \parbox{\mydecisioncolwidth}{Situation recognition} &
      Ontology &
      \iconcrossmark{} &
      \iconcrossmark{} &
      \iconcheckmark{} &
      \iconcheckmark{} &
      \iconcrossmark{} &
      \parbox{\myusecolwidth}{Video sequence\\from one person}\\
    & \parbox{\mypubcolwidth}{\citeauthor{NishiyamaEtAl2011}\\ (\cite*{NishiyamaEtAl2011})} &
      Predefined situations &
      Influence diagram &
      Bayesian model &
      \iconcrossmark{} &
      \iconcrossmark{} &
      \iconcrossmark{} &
      \iconcrossmark{} &
      -- &
      Synthetic data \\
    & \parbox{\mypubcolwidth}{\citeauthor{ChahuaraEtAl2017}\\ (\cite*{ChahuaraEtAl2017})} &
      \parbox{\myctxtcolwidth}{Location, activity, agitation,\\predefined situations} &
      \parbox{\mydecisioncolwidth}{MLN based\\influence diagram} &
      OWL2 and MLN &
      \iconcheckmark{} &
      \iconcrossmark{} &
      \iconcheckmark{} &
      \iconcheckmark{} &
      \iconcheckmark{} &
      37 naive users\\%
    \midrule{}
    \multirow{4}{=}[-3pt]{\centering\rotatebox[origin=c]{90}{Reinforcement}} &
      \parbox{\mypubcolwidth}{\citeauthor{Mozer2005}\\ (\cite*{Mozer2005})} &
      Location, activity &
      \parbox{\mydecisioncolwidth}{Reinforcement learning} &
      Ontology &
      \iconcrossmark{} &
      \iconcheckmark{} &
      \iconcheckmark{} &
      \iconcheckmark{} &
      \iconcheckmark{} &
      Daily life use\\
    & \parbox{\mypubcolwidth}{\citeauthor{KaramiEtAl2016}\\ (\cite*{KaramiEtAl2016})} &
      Activity &
      \parbox{\mydecisioncolwidth}{Reinforcement learning} &
      Ontology &
      \iconcrossmark{} &
      \iconcheckmark{} &
      \iconcrossmark{} &
      \iconcheckmark{} &
      \iconcheckmark{} &
      Synthetic data\\
    & \parbox{\mypubcolwidth}{\citeauthor{MnihEtAl2015}\\ (\cite*{MnihEtAl2015})} &
      None &
      \parbox{\mydecisioncolwidth}{Deep reinforcement\\learning} &
      Graphical &
      \iconcrossmark{} &
      \iconcheckmark{} &
      \iconcheckmark{} &
      \iconcrossmark{} &
      \iconcrossmark{} &
      Game simulation\\%
    \addlinespace
    \cmidrule(l){2-11}
    & \textbf{This study} &
      None &
      \parbox{\mydecisioncolwidth}{Deep reinforcement\\learning} &
      Graphical &
      \iconcheckmark{} &
      \iconcheckmark{} &
      \iconcheckmark{} &
      \iconcheckmark{} &
      \iconcheckmark{} &
      15 naive users\\
    \bottomrule
  \end{tabular}
  \hfill
\end{scriptsize}
\end{sidewaystable*}
\typeout{Landscape context table... END^^J}

\autoref{tab:related_works:sumup} lists the main attributes of the papers considered in this state of the art.
It shows the variety of models that have been applied for decision-making in contrast to the lack of experimental real-time tests in real environments.
Even when several approaches have been proposed to handle knowledge representation, inference or adaption, most of these works have been applied to a particular problem of context awareness, such as activity recognition, and did not consider the problem as a decision-making one.
In this work, we propose an approach using deep-learning to learn the features/concepts that comes into play during the decision at the same time as the decision model.
This learning is performed using the reinforcement learning paradigm so that on-line adaptation is made possible.
Furthermore, we propose an original knowledge presentation, raw image, to benefit from the recent improvement in image processing using deep-learning.
The remaining of this paper presents the method and its application to decision-making in a smart-home.
\clearpage{}
  \clearpage{}\section{Methods}
\label{sec:methods}

Typical smart-homes considered in the study are the ones that permit voice based interaction.
Voice-User Interfaces (VUIs) in domestic environments recently gained interest in the speech processing community.
The rising number of smart-home projects that consider Automatic Speech Recognition (ASR) in their design \parencite{IstrateEtAl2008,FilhoEtAl2010,LecouteuxEtAl2011,ChristensenEtAl2013} exemplifies this trend.
Adapted to homes containing multiple rooms which are fitted with sensors and actuators — such as infra-red presence detectors, consumption meters, multimedia servers, etc. — this kind of interface is particularly suitable to people in loss of autonomy \parencite{PortetEtAl2013,VacherEtAl2015}.
Smart-homes aim at providing daily living context-aware decision using the perception of the situation of the user, as previously illustrated in \autoref{sen:one}.

\begin{figure}
  \includesvg[\columnwidth]{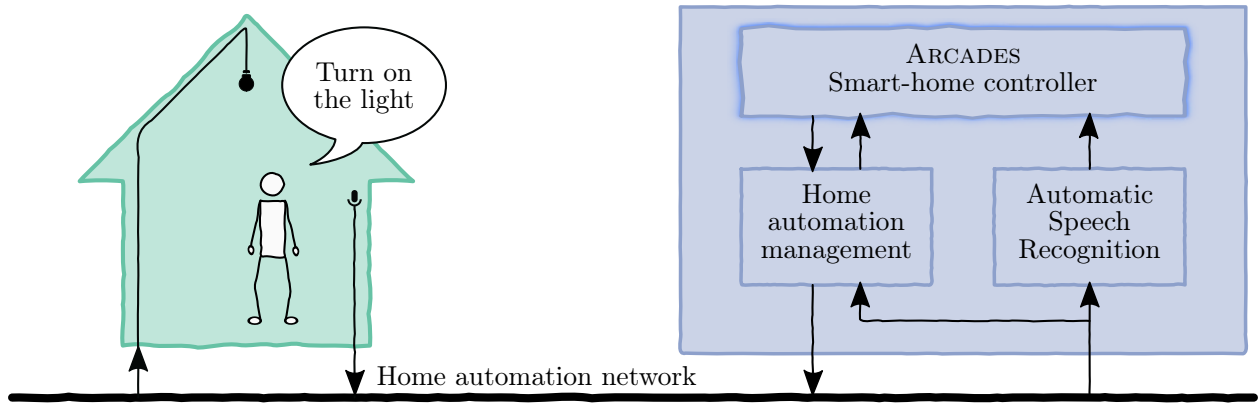}{smarthome}
  \caption[A typical smart-home system]{A typical smart-home system equipped with a voice-user interface: use case of a user asking to turn on the light.}
  \label{fig:methods:smarthome}
\end{figure}

The interactions between the user, her home and the control system is depicted in \autoref{fig:methods:smarthome}.
An underlying home automation network, composed of typical home automation sensors and actuators (switches, lights, blinds, etc.) and several microphones per room allows a controller to interact with the home and the user.

As emphasized in \autoref{sec:introduction}, the main challenges in such a context are to be able to handle heterogeneous information in a unique representation space and to adapt the decision-making to an evolving environment.
Thus, the remaining of this section focuses on the description of \ARCADES{}, the Adaptive Reinforced Context-Aware Deep Decision System, and its components.
We start with the reinforced adaptation loop (\autoref{sec:methods:RL}), then the data representation (\autoref{sec:methods:images}) and finally the decision model acquired using deep learning (\autoref{sec:methods:DL}).
Last section (\autoref{sec:methods:learning}) explains how all these components interact with each other to allow the system to learn.

\subsection{Adaptation of the decision model through reinforcement learning}
\label{sec:methods:RL}

\begin{figure}
  \centering
  \includesvg[0.75\columnwidth]{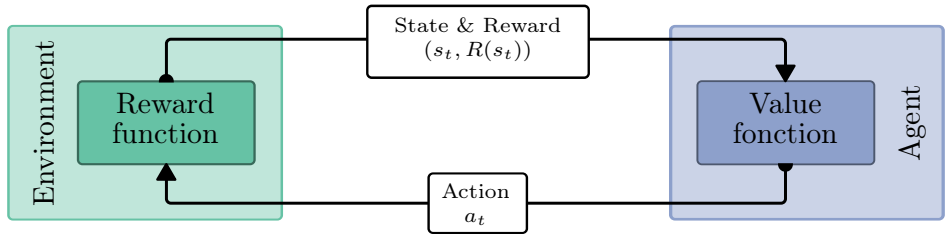}{systemComponents}
  \caption{The main components of \ARCADES{} and their interactions}
  \label{fig:methods:components}
\end{figure}

A Reinforcement Learning (RL) task relies on the Markov Decision Process (MDP) formalism which considers two components: the \emph{environment} and the \emph{agent}.
\autoref{fig:methods:components} shows the interactions between each component: the agent (here the decision-making system) is fed with the current environment state $s_t \in \mathcal{S}$.
It chooses an action $a_t \in \mathcal{A}$ based on this state.
The chosen action affects the environment (here the house and the user) whose state changes (from $s_t$ to $s_{t+1}$) and which generates a reward.
This reward $R(s_t) = r_t$ can be positive if the action is appropriate, negative if the action is not adequate, or null.
Based on this reward, the agent adapts its decision policy.
In \autoref{sen:one}, the voice command and context of the interaction compose the state, while the decision made by the system is the action to execute.
Then the user will be able to evaluate/reward this decision by different means not discussed here.

In a fully qualified MDP, dynamic programming \parencite{Bellman1957} is used to compute the optimal policy.
However, it cannot be used in real world problem due to missing information about the environment model and the reward function.
In that case, \emph{$Q$-Learning} \parencite{Watkins1989,SuttonEtAl2017} can be used to compute and update a policy which will converge toward the optimal one.

In $Q$-Learning, a function, named \emph{$Q$-function}, associates a value, named \emph{$Q$-value}, to each possible pair of state-action, named $Q$-state: $Q : \mathcal{S} \times \mathcal{A} \to \mathbb{R}$, where $\mathcal{S}$ is the set of possible states of the environment, and $\mathcal{A}$ is the set of possible actions to perform.
Beginning with an initial $Q$-function ($Q_0$), it can be updated after each interaction $\langle s_t, a_t, r_t, s_{t+1} \rangle$, following \autoref{eq:methods:qLearning} (where $\gamma$ is the discounting factor of the reward function and $\alpha$ is a learning rate).
This updating rule has been proven to converge toward the actual $Q$-value of each $Q$-state, allowing to compute the optimal policy to solve the task.

\begin{equation}
Q_{k+1}(s_t, a_t) = \alpha \big( r_t + \gamma max_a Q_k(s_{t+1}, a) \big) + (1-\alpha) Q_k(s_t, a_t)
\label{eq:methods:qLearning}
\end{equation}

The following sections describe how the states of the environment can be acquired, and the model used for the $Q$-function, before presenting how both components interact during the learning process.

\subsection{A common representation for heterogeneous information: a raw image approach}
\label{sec:methods:images}

One challenge in reasoning with raw data is to define a referential space where all kinds of information can be represented and manipulated.
Indeed, real smart-home data are multi-modal \parencite{VacherEtAl2014}: from continuous time series, such as the overall water consumption or temperature, to event based semantic information such as voice commands or the state change of a device.
Hence, finding a representation that can accommodate low-level sensor events (presence in a room) and high-level events (user current activity) is a problem that has not found a consensual solution.
Here, we are proposing the use of a \textbf{graphical representation}, projecting any information on a two-dimensional map of the smart-home any time the environment state changes.
Then, we will rely on the ability of Convolutional Neural Networks (CNN) to build relevant features from this raw representation.
This approach provides some advantages listed below.
\begin{enumerate}
	 \item The images generated can be used by the user to monitor what the system sees or understands. 
	\item The image size is only loosely linked to the number or type of sensors (adding, removing, or changing some sensors will not change the size of the image), hence images accommodate a variable number of data sources (sensors can come and go) a property that classical input vector machine learning approaches do not \emph{naturally} handle.
	\item Images \emph{naturally} convey spatial information (here an \emph{a priory} information about the home room organisation).
	\item The high redundancy of sensor values can be efficiently used by deep learning.
\end{enumerate}

\begin{figure}
  \centering
  \includegraphics[width=0.8\columnwidth]{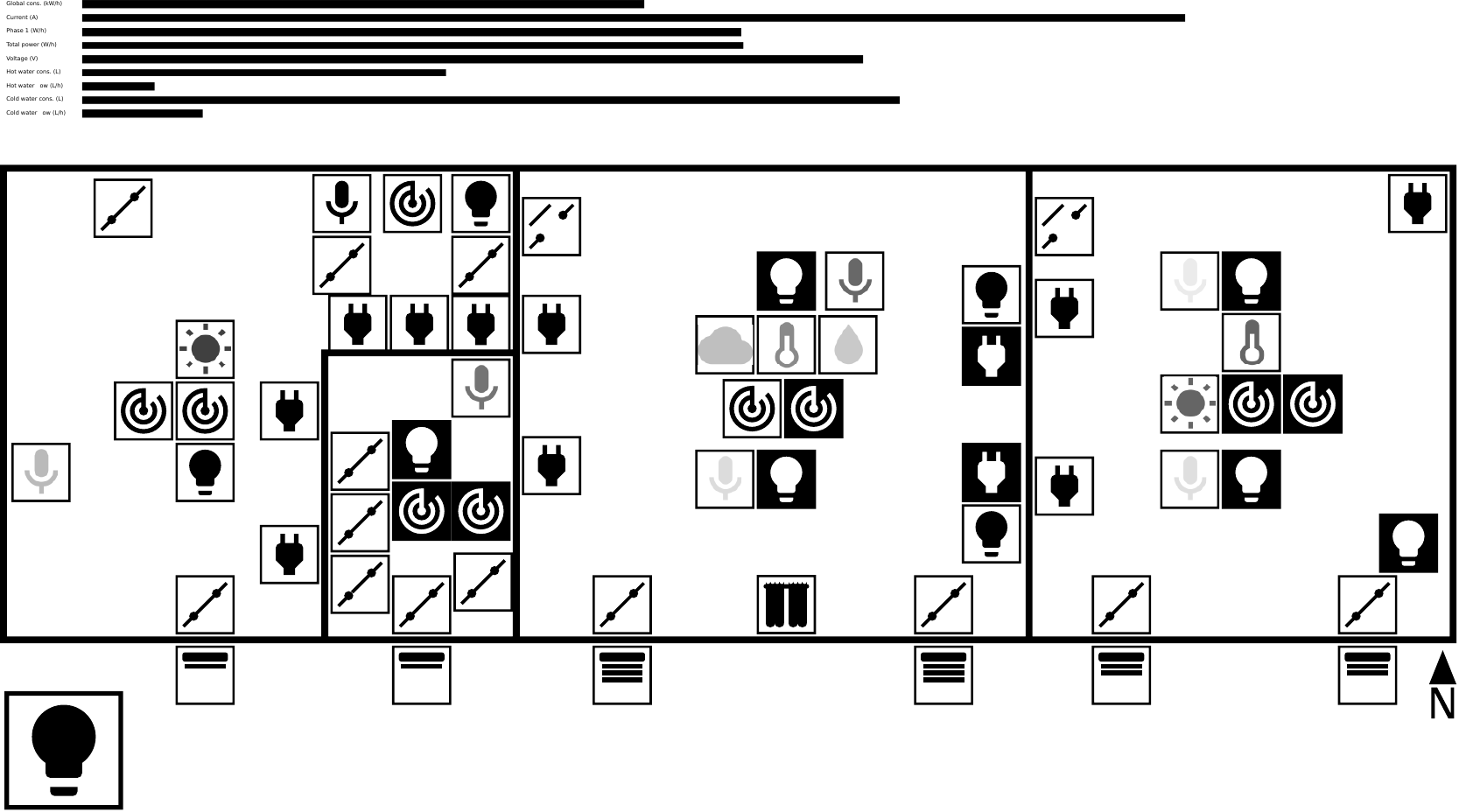}
  \caption{Examples of image generated for the agent input. From left to right : kitchen, bathroom, bedroom and study.}
  \label{fig:methods:image}
\end{figure}

Example of generated image for the smart-home considered in the study are provided in \autoref{fig:methods:image}.
In this image, as in the \autoref{sen:one}, the user asks to turn on the light while she is washing up dishes.
The smart home walls are represented by black lines while icons represent raw sensor data.
For instance usage of light is represented by a black lamp on white background while a turned off lamp is a white lamp on dark background.
The highest part presents some gauges for electricity and water consumption.
The big icon at the bottom left represents the uttered command.
Finally, other icons match the last known state of each information provider (sensor or inference system) of the home, using a different icon for different information type.
It exists a dedicated icon for various kinds of information such as:
\begin{itemize}
  \item Binary sensors symbols: presence detector, door/window detector, blind state, light state, etc.;
  \item Continuous sensors symbols: light level, sound level, temperature, humidity;
  \item Continuous sensors gauges: electricity consumption, water consumption, etc.;
  \item High level information symbols: uttered command.
\end{itemize}

The image and icons have been chosen to maximise the contrast and redundancy for the input of a CNN, to be understandable by human and to be easily generated on the fly.
In fact, the implementation relies on the SVG\footnote{Scalable Vector Graphics (SVG) is a widely-deployed, royalty-free, XML-based vector image format developed and maintained by the World Wide Web Consortium (W3C) (\url{https://www.w3.org/Graphics/SVG/})} technology, to make it possible to update the image on-line to reflect the state of the smart-home.

\subsection{Decision making model through deep-learning}
\label{sec:methods:DL}

As explained in \autoref{sec:methods:RL}, the decision policy rely on the $Q$-function of the agent (the simplest policy being: \enquote{Given $s$, choose the action with the highest $Q$-value}) and we are going to present two ways to represent it.

Classical implementations choose a tabular representation. A two-dimensional array, indexed by states in one dimension and by actions in the other, records the $Q$-values for each $Q$-state.
However, this approach does not scale to big or continuous sets of states or actions and cannot capture redundancy in states.

Another recent implementation relies on a deep neural network to model the $Q$-function. This representation has been introduced by \textcite{MnihEtAl2015} who introduced the Deep Q-Network (DQN), which has been successfully tested on decision-making task related to Atari2600 games (cf. related work \autoref{sec:related_works}). 
Although, their application was different from our, the voice command problem can be seen as a general problem of decision-making in a changing environment.
We thus adapted the approach to smart-home decision-making.

The decision model is based on a Convolutional Neural Network (CNN) to process the images followed by a fully connected neural network to choose the best action to perform.
\autoref{fig:methods:DL:architecture} presents the architecture of our $Q$-network.
While not discussed here, different architectures has been tested, taking into account base architecture proposed by \textcite{MnihEtAl2015}, technical limitation like memory consumption as well as some \textit{a priori} knowledge like the \SI{6}{px} based grid of the images.
Finally, the neural network is composed of four \emph{convolutional layers}, reducing a  $256 \times 256$ input image to features maps of sizes: $16 \times 64 \times 64$, $32 \times 21 \times 21$, $64 \times 10 \times 10$, and $64 \times 8 \times 8$.
The features maps of the last layer are then rearranged in a vector of dimension $64 \times 8 \times 8 = 4096$.
This vector passes through \emph{fully connected linear layers} of width $512$ and $33$ which is the number of possible actions.
A \emph{rectified linear unit} separates each layer.

\begin{figure*}
  \includesvg[\textwidth]{./assets/images/methods}{networkArchitecture}
  \caption[Architecture of the $Q$-network]{
  Architecture of the $Q$-network.\\
  $C_n[f+p/k]$ denotes a convolution layer (using $n$ filters of size $f$, with a padding of $p$, and using a stride of $k$) and $FC_n$ a fully connected linear layer (of width $n$).}
  \label{fig:methods:DL:architecture}
\end{figure*}

Thus, this model provides a way to process the generated images presented in \autoref{sec:methods:images}.
After some preprocessing (scaling and reducing the depth of the image), the image is passed forward in the $Q$-network.
The network extracts relevant features used to classify it in the range of possible actions.
It provides, as output, the $Q$-values for each action given the input state (and thus can be seen as a function $Q\text{-network} : [0,255]^{256 \times 256} \to \mathbb{R}^{33}$).
Then, \autoref{eq:methods:qLearning} detailed earlier can be used as a loss function to update the weights of the $Q$-network, as in typical deep learning approaches.

While the work from \citet{MnihEtAl2015} is not the first attempt to use neural network as $Q$-function \parencite{TsitsiklisEtAl1997}, it seems to be the first successful one.
This is due to different optimization techniques as the use of a target network, experience replay \parencite{Lin1993}, mini-batch learning or RMSProp gradient optimization \parencite{Hinton2015}.
All these techniques have been kept in our implementation and the reader is referred to the original paper \parencite{MnihEtAl2015} for detailed explanations.

\subsection{Learning method}
\label{sec:methods:learning}

\begin{algorithm}
  \begin{algorithmic}
    \Require $epoch = 0$
    \While{$epoch < maxEpoch$}
    \State $step \gets 0$
    \Repeat
    \State $step \gets step + 1$
    \State obs $\gets$ \textbf{environment}.\Call{getObservableState}{}
    \State \textbf{agent}.\Call{integrateObservations}{obs}
    \State action $\gets$ \textbf{agent}.\Call{getAction}{}
    \State \textbf{environment}.\Call{performAction}{action}
    \State reward $\gets$ \textbf{environment}.\Call{getReward}{}
    \State \textbf{agent}.\Call{giveReward}{reward}
    \Until {$step > maxSteps$}
    \State \textbf{agent}.\Call{learn}{}
    \State $epoch \gets epoch + 1$
    \EndWhile
  \end{algorithmic}
  \caption[Interaction logic]{Logic used to allow the agent and the environment to interact.}
  \label{alg:methods:experiment}
\end{algorithm}
The overall learning of the decision-making module (the agent) is decomposed into two phases: a \emph{pre-training phase} (i.e., initialization of the agent), during which the model is learned from scratch and an \emph{adaptation phase} during which the model is tuned toward the specific inhabitant and environment.
In practice, the pre-training phase is performed off-line and is mostly devoted to the learning of the CNN and the subsequent neural network.
It is well-known that CNN needs a high amount of data that is why, in our work, we used artificially generated data to feed the CNN with raw input to learn the CNN weights.
In the adaptation phase the CNN is not supposed to evolve but the last part of the neural network (i.e., the decision part) is biased towards the adaptation data.
This data can be either acquired on-line or performed on corpus if available.
In any of these phases, the reinforcement learning method is the same and is described in the \autoref{alg:methods:experiment}.
Starting from an initial agent, the system makes $maxSteps$ interactions, in which the observable state of the environment is passed to the agent, before it chooses an action applied back to the environment which releases a reward.
All these interactions are recorded by the agent which uses them during a learning procedure.
This training epoch (interactions plus training) is realized a number $maxEpoch$ of times.

\begin{algorithm}
  \begin{algorithmic}
    \Function{performAction}{predictedAction}
    \If {$predictedAction \text{ is } trueAction \lor$ \\ $\quad numberOfTries > triesThreshold$}
    \State \Call{moveToNextDataSample}{}
    \State $numberOfTries \gets 0$
    \Else
    \State $numberOfTries \gets numberOfTries + 1$
    \EndIf
    \EndFunction
  \end{algorithmic}
  \caption[Simulation of the environment evolution]{
    Function used to simulate the evolution of the environment for synthetic data.}
  \label{alg:methods:environment:performaction}
\end{algorithm}

Since during the pre-training phase the agent explores artificially created data, the environment is not able to evolve consequently to a wrong action.
Thus, to multiply the number of situations and explore wrong decisions, the function described in \autoref{alg:methods:environment:performaction} simulates the environment evolution for synthetic data.
In this function, as long as the action chosen is not the correct one, up to $triesThreshold$ actions can be successively tried to simulate the patience of the user, repeating the same command until the system eventually behaves correctly.

Similarly, all the experiments presented in this work are realised on a previously recorded corpus.
Thus, it is not possible for the user to interactively generate a reward.
We defined a simple reward function able to simulate this reward, based on a ground truth (obtained from annotations of the corpus) and the predicted action: \emph{if the chosen action is correct emit \num{+1}, else emit \num{-1}}.

\begin{algorithm}
  \begin{algorithmic}
    \Require{$\epsilon \in [0; 1]$}
    \Function{getAction}{}
    \State $x \sim \mathcal{U}([0, 1])$
    \If{$x \le \epsilon$}
    \State action $\sim \mathcal{U}(\mathcal{A})$ \Comment{Random action}
    \Else{}
    \State $Q$-values $ \gets $ $Q$-Network.\Call{forward}{state}
    \State action $\gets \argmax (Q\text{-values})$
    \EndIf{}
    \State \textbf{return} action 
    \EndFunction{}
  \end{algorithmic}
  \caption{Agent policy}
  \label{alg:methods:agent:policy}
\end{algorithm}
Fetching the action to perform requires the agent to execute its policy described in \autoref{alg:methods:agent:policy}.
We use a near greedy policy named $\epsilon$-greedy policy \parencite{SuttonEtAl2017}.
Given $\epsilon \in [0;1]$, this policy returns a random action with probability $\epsilon$ or the action with the highest $Q$-value given the current state.
To compute the $Q$-value of each action, the agent uses its approximated $Q$-function, modeled by the deep neural network $Q$-Network.

\begin{algorithm*}
  \begin{algorithmic}
    \Function{learn}{}
    \State $\mathcal{B} \gets$ \Call{getSamples}{$minibatchSize$} \Comment{Each sample is $\langle s_{t}, a_{t}, r_{t}, s_{t+1} \rangle$}
    \State $\mathbf{Q}$\textbf{-values} $ \gets
    \big($ \textbf{$Q$-Network}.\Call{forward}{
      $\mathbf{\mathcal{B}_{s_t}}$
    }$ \big)
    \big[\mathbf{\mathcal{B}_{a_t}}\big]$ \Comment{Vector of $Q$-values of each $Q$-state in $\mathcal{B}$}
    \State \Comment{(0 for all $Q$-states not in the batch).}
    \State \textbf{targets-values} $\gets
    \mathbf{\mathcal{B}_{r_t}} + \gamma \max\big($
    \textbf{target-$Q$}.\Call{forward}{
      $\mathbf{\mathcal{B}_{s_{t+1}}}$
    } $\big)$
    \State \Comment{Target values computed with the target network}
    \State $\mathbf{loss} \gets $ \textbf{targets-values} $-\ \mathbf{Q}$\textbf{-values}
    \State $\mathbf{gradient} \gets$
    \textbf{$Q$-Network}.\Call{backward}{
      $\mathbf{\mathcal{B}_{s_t}}, \mathbf{loss}$
    }
    \State $\mathbf{updatedGradient} \gets$ \Call{RMSprop}{$\mathbf{gradient}$}
    \State \textbf{$Q$-Network}.$parameters \gets$
    \textbf{$Q$-Network}.$parameters + \mathbf{updatedGradient}$
    \EndFunction
  \end{algorithmic}
  \caption[Agent learning procedure]{
    Agent learning procedure\\
    Notation: $\mathbf{\mathcal{B}_{s_t}}$ is the vector composed of the $s_t$ component of each sample.}
  \label{alg:methods:agent:learning}
\end{algorithm*}

The learning of the agent follows the work of \textcite{MnihEtAl2015}.
In particular, to estimate the $Q$-value the learning does not use the latest experienced interactions.
Instead, each interaction $\langle s_t, a_t, r_t, s_{t+1} \rangle$ is recorded in a pool.
Then, a set $\mathcal{B}$ of interactions is randomly drawn from this pool and used as a \mbox{(mini-)batch} to train the network.
An updated $Q$-value, more accurate than the current one (taking account of the reward provided by the user) and named target-value, is computed for each $Q$-state involved in $\mathcal{B}$ according to: $\text{target-value} = r_t + \gamma \max_a Q(s_{t+1}, a)$.
The difference between current $Q$-values and target-values is used as a loss and back-propagated through the \textbf{$Q$-Network} to compute the gradient.
The gradient obtained is optimized using the RMSprop optimization algorithms \parencite{Hinton2015} before being applied to the weights of the neural network.
This whole process is described in \autoref{alg:methods:agent:learning}.
\clearpage{}
  \clearpage{}\section{Experiment}
\label{sec:experiment}

\begin{figure}
  \centering{}
  \includesvg[\columnwidth]{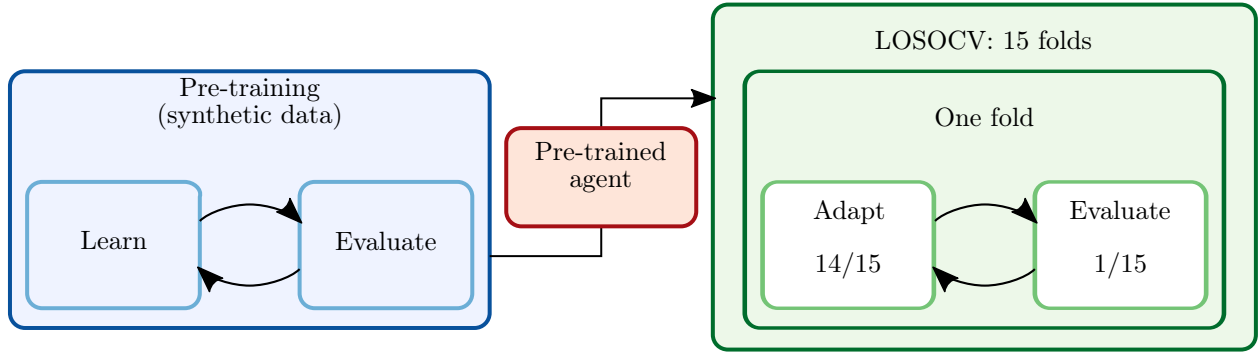}{expe_steps}
  \caption[Experiment protocol]{
    Experiment protocol. After pre-training with synthetic data, the agent is adapted with data from 14 of 15 subjects and evaluated on the left out participant's data.}
  \label{fig:expe:steps}
\end{figure}

The experiment conducted in this study was decomposed in two phases as illustrated in \autoref{fig:expe:steps}.
The first phase, the \textbf{pre-training phase}, uses a generated corpus to train an agent and a similar corpus to test it.
The \textbf{validation phase} then, performs a cross-validation on a real corpus.
As data recorded for each participant is comparable, a \textbf{Leave One Subject Out Cross-Validation} (LOSOCV) method is used to avoid bias.
It uses fourteen of fifteen participants' data for adapting and tests \ARCADES{} with the left out participant's data.

\subsection{Input data}
\label{sec:methods:data}

\subsubsection{Sweet-Home Corpus}

\begin{figure}
  \centering{}
  \includegraphics[width=0.75\columnwidth]{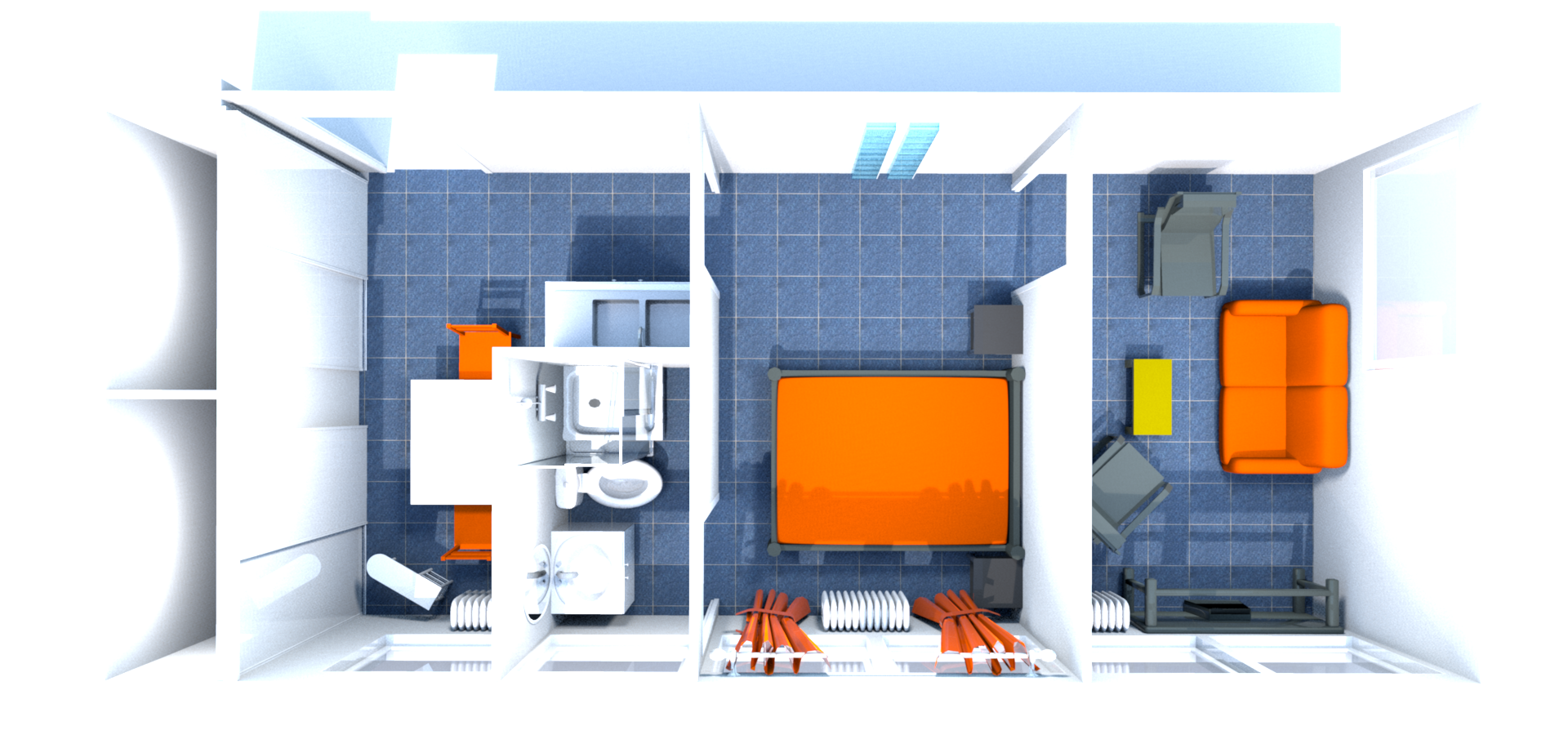}
  \caption{The Domus smart-home}
  \label{fig:experiment:domus}
\end{figure}

The data corpus was recorded in the \textsc{Domus} smart-home designed by Laboratoire d’Informatique de Grenoble (LIG) \parencite{GallissotEtAl2013}.
This \SI{30}{\square\meter} home includes a bathroom, a kitchen, a bedroom, and a study as show in \autoref{fig:experiment:domus}.
All these rooms are equipped with sensors and actuators such as infrared motion sensors, tactil sensors, temperature sensors, power meter, etc.
In addition, seven microphones in the ceiling capture the audio.
The apartment is fully usable and can accommodate a resident for several days.
More than 150 devices are managed in the apartment to provide different services (e.g. light, opening/closing shutters, media management, etc.).

To collect decision data in context, several participants were recruited to play scenarios of the daily life in the apartment.
While enacting daily life scenarios, participants had to issue voice commands to activate the actuators in the smart home.
The objective of this experiment was to test a smart controller in real situations corresponding to voice commands spoken by the user.
This study considered the following situations: to cleanup the apartment; to prepare and eat a meal; to converse via video conference; to do leisure activities (reading); to take a nap.

To guide the participants, the grammar voice command was provided with a scenario of everyday life.
This scenario was designed to last about 45 minutes, however, there was no constraint on the execution time.
Each participant received a list of actions to make and voice commands to utter that was interpreted by an intelligent system acting accordingly.
Each participant had to use a voice command, repeating up to 3 times in case of system failure, before a Wizard of Oz was used.

A total of 15 people (9 women, 6 men) participated in the experiment to record 11 hours of data.
The average age of participants was $38 \pm 13.6$ ($19$ -- $62$).
All experiments were video recorded but only for annotation purposes.
This data set is part of the publicly available Sweet-Home corpus \parencite{VacherEtAl2014}.

\subsubsection{State and action sets}

The state space can be seen from different points of view.
If the state space is represented only by the location and activity of the user (i.e., classical context variables), the number of states is 324 \parencite{BrenonEtAl2016}. But these contextual information necessitate inference models and our aim is to directly make decision from raw data.
When state space is represented by raw data issued by the 81 available sensors (49 binary sensors and 32 continuous sensors which can take values in different ranges -- from $[0;100]$ up to $[0;2^{32}-1]$), this leads to a big number (more than $10^{12}$) of possible states. 
This shows again that decision making from raw data cannot be performed using explicit representation and that approximation of the kind presented here with neural network is the only way to make this decision making tractable. 

\begin{table}[b!]
  \vspace*{-1em}
  \centering{}
  \caption{Available actions for the agent to act on the environment.}
  \label{tab:expe:sets:actions}
  \begin{tabular}{p{0.17\columnwidth}p{0.75\columnwidth}}
    \toprule{}
    Command & Actuator devices\\%
    \midrule{}
    turn on the light & all lights in the kitchen, light above the sink in the kitchen, ceiling light of the kitchen, all lights in the bedroom, bedside lamp of the bedroom, ceiling light of the bedroom, ceiling light of the study\\%
    \midrule{}
    turn off the light & all lights in the kitchen, light above the sink in the kitchen, ceiling light of the kitchen, all lights in the bedroom, bedside lamp of the bedroom, ceiling light of the bedroom, ceiling light of the study\\%
    \midrule{}
    turn on the radio & bedroom radio\\%
    turn off the radio & bedroom radio\\%
    \midrule{}
    open the blinds & kitchen blinds, bedroom blinds, study blinds\\%
    close the blinds & kitchen blinds, bedroom blinds, study blinds\\%
    \midrule{}
    open the curtains & bedroom curtains\\%
    close the curtains & bedroom curtains\\%
    \midrule{}
    give time & kitchen speakers, bedroom speakers, study speakers\\%
    give temperature & kitchen speakers, bedroom speakers, study speakers\\%
    \midrule{}
    call emergency & study phone\\%
    call a parent & study phone\\%
    \midrule{}
    do nothing & nowhere\\%
    \bottomrule{}
  \end{tabular}
  \vspace*{-2em}
\end{table}

The action set depends on the number of actuators and commands concerning them.
Even if some actions were never realized during the corpus record, we chose to handle a large set of actions.
Thus, any device used at least once has been incorporated, with all its possible commands even if some of them are not used.
This ended up with a set of 33 possible actions listed in \autoref{tab:expe:sets:actions}.

\subsubsection{Generated corpus}

From a deep-learning point of view, the Sweet-Home corpus does not contain enough relevant samples of voice commands.
Furthermore, it contains unpredictable samples or missing data, which is not advisable for a first proof of concept.
As a workaround, a data generation system was developed based on a few generation rules.
This system can be tuned to generate deterministic data (which always associate the same output to the same input), restricted data (focusing only on most significant events), etc.

The generated corpus follows the same format as the real raw data corpus.
Thus, the subsequent translation of the annotations to SVG image (cf. \autoref{sec:methods:images}) can be applied seemingly to the real or generated data.
The generation system producing samples on-demand, this corpus is theoretically of infinite size, a new sample always being available if necessary.

The corpus is generated to use the ability of the deep learning to build its own features from raw data.
Thus, the generation consisted in generating annotations for the voice command and the expected action as well as simulating values for the raw sensor data.
To achieve this, we generate an \enquote{annotated} state, similar to the ones used in our preliminary study \parencite{BrenonEtAl2016}, which is converted using logical rules.
For example, if the annotated location is the kitchen, then the presence detector sensors of the kitchen are constrained to be active.
In the meantime, the bedroom being the room next to the kitchen, its presence detector sensors are constrained to be active with a defined probability.
This mechanism is applied for all relevant sensors, while random values are applied to irrelevant ones.
The real Sweet-Home corpus is a set of CSV files, one for each sensor plus two for the uttered command annotation and the expected action (the target of the learning).
These files contain the value for the data source for each time-stamp which can be parsed to generate a data structure similar to the one generated by our generation system.

\subsection{Experiment: learning on raw data}
\label{sec:expe:2}

First attempts to train a model from images generated with raw data lead to poor performances (oscillations, long convergence time, etc.).
To address this problem, we choose to investigate the hyper-parameter choices, conducting an exhaustive search for three of them which are \targetq{}, \minibatchsize{} and \updatefreq{}.
Result of this research is summed up in \ref{sec:appendix:hyperoptim}.
These parameters are used by the agent, \minibatchsize{} and \updatefreq{} being directly related to the learning procedure described in \autoref{alg:methods:agent:learning}.
The former is the number of samples used for the training, that is to say the size of $\mathcal{B}$.
The latter is the number of steps done before calling the \textsc{learn} procedure.
As explained, the learning does not happen after each interaction but after each \updatefreq{} interactions.
The last parameter, \targetq{}, is related to the target network trick introduced by \textcite{MnihEtAl2015}, and the whole details of its use is out of the scope of this paper.
However, these three parameters are tightly linked.
A high \targetq{} and \minibatchsize{} must improve the overall stability of the $Q$-network, but a high \minibatchsize{} leads to a longer training time.
Thus, increasing the \updatefreq{} must reduce this learning time.

While the data streams of synthetic generated corpus are perfectly synchronized, this is not the case for the real data.
Despite our efforts to synchronize sensor streams and annotations, the real smart-home corpus does not provide a high quality of synchronization.
To overcome this problem, we choose to provide to the $Q$-network a wider window of the last 3 seconds of the state of the environment.
However, this mechanism generate many samples where no action is required by the agent (2 samples of no action for one sample of any of the 32 other actions), and thus heavily unbalance our classes.
To overcome this, the reward function was updated to heavily penalize an error about this majority class.
Instead of receiving a reward of $(-1)$, the agent receives a reward of $(-192)$ if it chooses to do nothing while an action is expected.

\begin{table}[!ht]
  \centering
  \caption{Parameters of the experiment.}
  \label{tab:expe:config:2}
  \small
  \begin{ArcadesConfig}*[][0.49\columnwidth]
    \caption{Pre-training}
    \gittag{synthetic-raw-full-1}
    \data{synthetic, raw}
    \totalsteps{\num{10e6}}
    \learningsteps{\num{100e3}}
    \evalsteps{\num{5000}}
  \end{ArcadesConfig}
  \begin{ArcadesConfig}*[][0.49\columnwidth]
    \caption{Validation}
    \gittag{real-raw-full-1}
    \data{real, raw}
    \totalsteps{\num{400e3}}
    \learningsteps{\num{4000}}
    \evalsteps{\num{5000}}
  \end{ArcadesConfig}
  \vspace*{-2em}
\end{table}

The parameters of this experiment are detailed in \autoref{tab:expe:config:2}.
The \emph{pre-training phase} used synthetic data generating images similar to the one presented in \autoref{fig:methods:image}.
The model was trained using up to \num{10e6} interactions, evaluating the system on \num{5000} independent interactions every \num{100e3} interactions.
Once a model is learned from scratch, the \emph{validation phase} adapts it to new data.
Each fold of the LOSOCV method runs the system for \num{400e3} interactions on the real data.
Test of \ARCADES{} happens every \num{4000} interactions, running \num{5000} monitored interactions.

To compare against a classical approach a {\bf baseline} system using a tabular Q-learning has been learned and tested by LOSOCV. However, since the classical approach uses an array to represent the state, this approach would not have been tractable if it was applied directly to the large amount of sensors values (cf.~\autoref{sec:methods:DL}). Hence, the baseline was trained on the ground truth humanly annotated location and activity labels. 
An except of the ground truth data is showed below. It shows that a voice command to open the blinds in the kitchen should be followed by the opening of all
the blinds of the kitchen. The second line is related to \autoref{sen:one}, when the light is asked to be turned on while cooking in the kitchen, the only possible action is to light on the lamp above the sink.
\begin{verbatim}
open - blind   kitchen  none  -> open  blind  kitchen
on   - light   kitchen  cook  -> on    light  kitchen - sink
\end{verbatim}
Thus this kind of ground truth dataset favours the baseline performances.

\autoref{tab:learning_results} summarizes the result of the learning. Even though the reinforcement learning objective is to maximise the reward, in this paper, standard classification measures such an F-Measure have been chosen. Indeed, the task can be seen as a classification one, classifying a sample (the environment state) between 33 classes (the actions). In our experiment, a correct classification is considered only when the chosen action is composed of the right action (e.g., turn on) on the right device (e.g., light) at the right place (e.g., study). Any other choice is a wrong classification. 

\begin{table}[!ht]
  \vspace*{-0.5em}
  \small
  \centering
  \caption{F-Measure of the pre-training and evaluation phases for the decision making task.}
  \label{tab:learning_results}
  \begin{tabular}{l|c|c}
	\toprule
	 & Pre-training F-M (\%) & Validation F-M (\%)\\
   &  synthetic data       &  real corpus \\
  \midrule
  Baseline\textsuperscript{1} &  NC  & 46.0 \\
  \ARCADES{}\textsuperscript{2}& 100 &  67.5 \\
  \bottomrule
  \end{tabular}

  \begin{footnotesize}  
    1. Learned and evaluated on humanly annotated data.\hspace{2em}2. Learned and evaluated on raw data.
  \end{footnotesize}
  \vspace*{-0.5em}
\end{table}

While the baseline tabular Q-learning only deals with humanly annotated data, it presents a low F-Measure of $46$\%. 
\ARCADES{} on raw sensor data does not lead to very high performances, but it demonstrates that a context aware decision-making model without an explicit representation of context is possible and lead to a very promising result of $67.5$\% of F-Measure. 

Detailed results of \ARCADES{} are presented \autoref{fig:experiment:eval:raw:pretrain} and \autoref{fig:experiment:eval:raw:real}. These are presented using the following metrics:
\begin{itemize}
  \item \emph{Reward per episode}: This value is the objective of the agent.
    It is the average sum of rewards obtain during each episode (an independent set of interactions).
    Given the reward function and the maximum length of an episode, the reward function cannot be greater than $1$.
  \item \emph{Precision, Recall, F1-Score}: standard performance measures for classification.
    \ARCADES{} can be seen as a classifier, classifying a sample (the environment state) between 33 classes (the actions).
\end{itemize}

First, it is interesting to note that the evaluation performances are better than the learning ones. This is explained with the exploration factor (the $\epsilon$ of the $\epsilon$-greedy policy), which is decreased from \num{0.99} to \num{0.5} during the first \num{200000} learning interactions (after that, 1 action over 2 is selected randomly) while it is set to \num{0} during the evaluation phase (greedy policy, no random decision).

Second, while being very long, the \emph{pre-training phase} finally reaches a very high plateau.
With more than \SI{4}~days and \SI{16}~hours of training \ARCADES{} acquires a coherent behavior as of the \num{8e6}\textsuperscript{th} interaction (about 100 days with one decision per second), as shown in \autoref{fig:experiment:eval:raw:pretrain}.
Then, during the \emph{validation phase} the system slightly adapts to the real data, starting with low performances, to finally reach more than \SI{65}{\percent} of F1-Score (\autoref{fig:experiment:eval:raw:real}).
Compared to the pre-training phase, this adaptation happens in a short time, comparable to the one obtained with the baseline system. 

\begin{figure}[p]
  \centering{}
  \arcadesExpeResults{%
    experiment:eval:raw:pretrain}{%
    \num{0.99}}{\num{100}}{\num{100}}{\num{100}}{%
    Pre-training}{%
    \emph{Pre-training} results}{%
    ./assets/gnuplot/eval/raw/synthetic}
  \bigskip{}
  
  \hrulefill{}
  \vspace{2\bigskipamount}
  
  \arcadesExpeResults{%
    experiment:eval:raw:real}{%
    \num{0,36}}{\num{68.91}}{\num{66.36}}{\num{67.49}}{%
    Validation}{%
    \emph{Validation} results}{%
    ./assets/gnuplot/eval/raw/real}
\end{figure}

\clearpage{}
  \clearpage{}\section{Complementary experiments}
\label{sec:complementary}

While in the case of a classical tabular $Q$-learning, the associative table can be interpreted, this is more complicated for the parameters of a neural network.
This is a recurrent grief addressed to neuronal models, and thus methods have been developed to better understand how they work \parencite{Karpathy2015}.
In this section, we present some complementary experiments aiming at understanding \ARCADES{} mechanisms.

In \autoref{sec:complementary:hardware} we test the ability of the system not to adapt to user preferences, but to changes in the sensor topology.
Then, we investigate, in \autoref{sec:complementary:tsne}, which features the neural network builds, and if they match the ones usually found in the literature.

\subsection{Adaptation to the hardware}
\label{sec:complementary:hardware}

One of the main goals of \ARCADES{} is to adapt itself to the user's preferences, but also to the sensors.
To verify this hypothesis, we made experiments in which some sensors disappear (because of faulty sensors or if the user uninstall them).
Four experiments were performed, based on a pre-trained (with synthetic raw data) model, in which different kinds of sensor were removed.

\begin{figure*}[p]
  \setlength{\abovecaptionskip}{0.33\baselineskip}
  \setlength{\belowcaptionskip}{0.33\baselineskip}
  \centering{}
  \begin{minipage}{0.99\columnwidth}
    \centering{}
    \includeplot[0.49\columnwidth]{./assets/gnuplot/complementary/faulty/PIR}{rewardPerEp}
    \includeplot[0.49\columnwidth]{./assets/gnuplot/complementary/faulty/PIR}{f1Score}
    \caption{Performances without motion sensors}
    \label{fig:complementary:faulty:PIR}
  \end{minipage}
  
  \hrulefill{}

  \begin{minipage}{0.99\columnwidth}
    \centering{}
    \includeplot[0.49\columnwidth]{./assets/gnuplot/complementary/faulty/sound}{rewardPerEp}
    \includeplot[0.49\columnwidth]{./assets/gnuplot/complementary/faulty/sound}{f1Score}
    \caption{Performances without acoustic noise}
    \label{fig:complementary:faulty:micro}
  \end{minipage}

  \hrulefill{}

  \begin{minipage}{0.99\columnwidth}
    \centering{}
    \includeplot[0.49\columnwidth]{./assets/gnuplot/complementary/faulty/switch}{rewardPerEp}
    \includeplot[0.49\columnwidth]{./assets/gnuplot/complementary/faulty/switch}{f1Score}
    \caption{Performances without door switches}
    \label{fig:complementary:faulty:switch}
  \end{minipage}

  \hrulefill{}

  \begin{minipage}{0.99\columnwidth}
    \centering{}
    \includeplot[0.49\columnwidth]{./assets/gnuplot/complementary/faulty/ASR}{rewardPerEp}
    \includeplot[0.49\columnwidth]{./assets/gnuplot/complementary/faulty/ASR}{f1Score}
    \caption{Performances without speech recognition}
    \label{fig:complementary:faulty:ASR}
  \end{minipage}
\end{figure*}

In the first experiment, the kitchen motion sensors were removed. \autoref{fig:complementary:faulty:PIR} shows that the performances are impacted but that the system quickly adapts its behavior to reach back its initial performances.

In the next experiments, the acoustic noise from the kitchen microphone was removed, then door switches for kitchen cupboard and fridge.
The latter are particularly active when the user tidy the home, and then we expected a dip of the performances.
However, for these two experiments, no impact can be measured (\autoref{fig:complementary:faulty:micro} and \autoref{fig:complementary:faulty:switch}) suggesting that the information obtained from these sensors are of little use to \ARCADES{}.

The last experiment simulates the break of the automatic speech recognition system.
While we could expect a drastic fall of the performances, near the random agent which would get about \SI{3}{\percent} of F1-Score, the system still acts not so badly, with an F1-Score oscillating around \SI{20 +- 10}{\percent}.
Noticing that the average reward is not so far below $(0)$ is another hint that its behavior is far from random.
One way to explain this, is that the system still gets images 2 seconds before an action is expected.
This information seems to be grabbed by the system during pre-training, and used to know when an action is expected (even if it is not the right one).
However, the oscillations of the score shows that \ARCADES{} is not able to infer the user intention only from the sensor values (i.e. from contextual information).

These results pushed us to ask which features the system does extract to make its decision.

\subsection{State features}
\label{sec:complementary:tsne}

During its forward pass through the network, the input image is transformed to extract relevant features, resulting in high-dimensional vectors.
One way to study these vectors is to project them in a two-dimensional space while preserving some inter-vectors relations such as the euclidean distance.
One of the most used projection method is the t-SNE (t-distributed Stochastic Neighbor Embedding) projection \parencite{MaatenEtAl2008} which perform very well to project vectors issued by a convolutional neural network.

In the following experiment, we provide some states (images) to a pre-trained model, while recording vectors after the last convolutional layer, and the one before the last fully connected layer.
These vectors are also linked to the information characterizing the input state (location, activity, uttered command, expected action, sensors activity).
Projecting the vectors provide a bi-dimensional coordinate $(0, 0) \leq (x, y) \leq (1, 1)$ where to represent any of the linked information.

\typeout{^^Jt-SNE...^^J}
\begin{figure*}
  \centering
  \setlength{\fboxsep}{0pt}

  \begin{subfigure}[b]{0.49\columnwidth}
    \centering
    \fbox{\includegraphics[width=0.99\columnwidth]{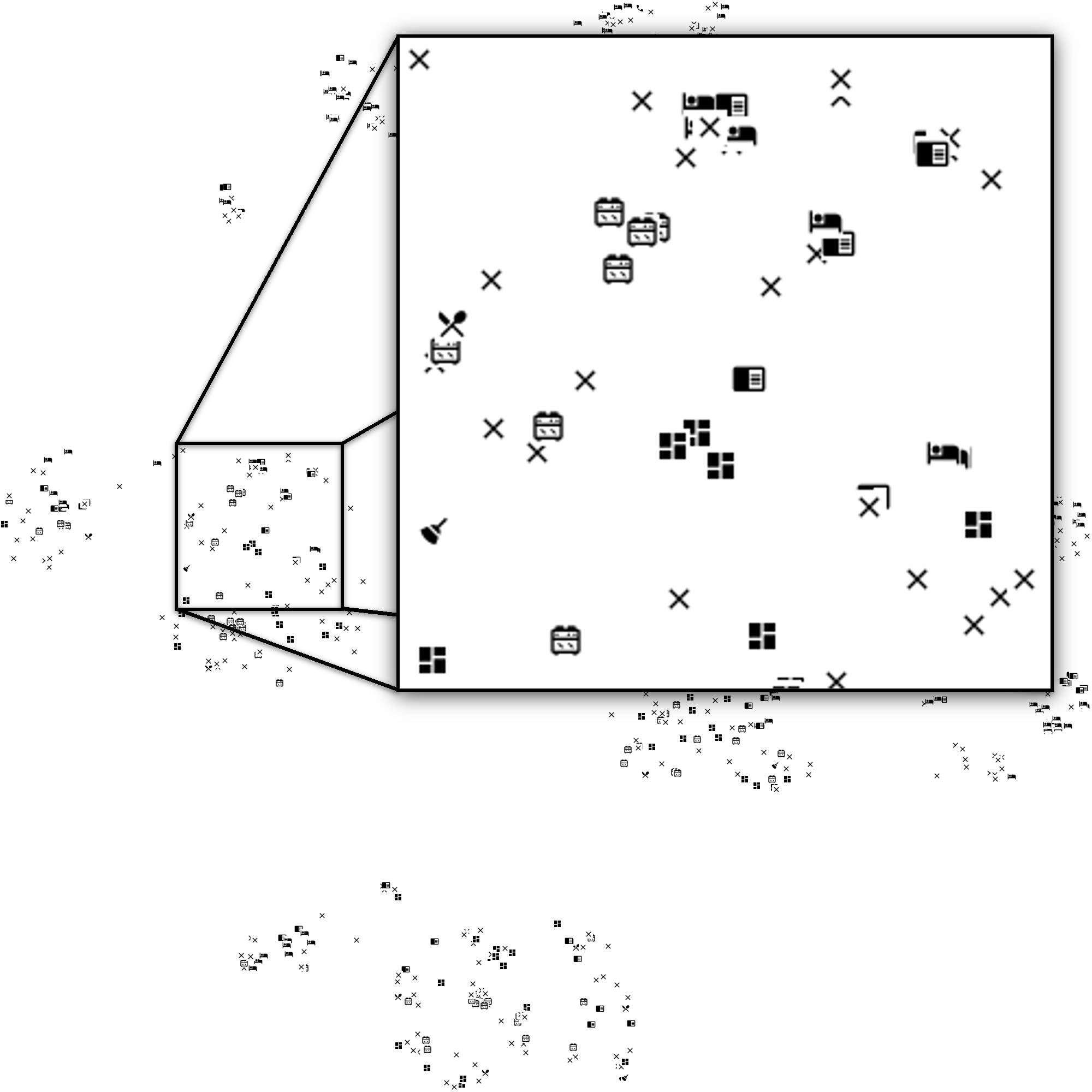}}
    \caption{Post-convolution: user activity}
    \label{fig:complementary:tsne:raw:conv:activity}
  \end{subfigure}
  \hfill{}
  \begin{subfigure}[b]{0.49\columnwidth}
    \centering
    \fbox{\includegraphics[width=0.99\columnwidth]{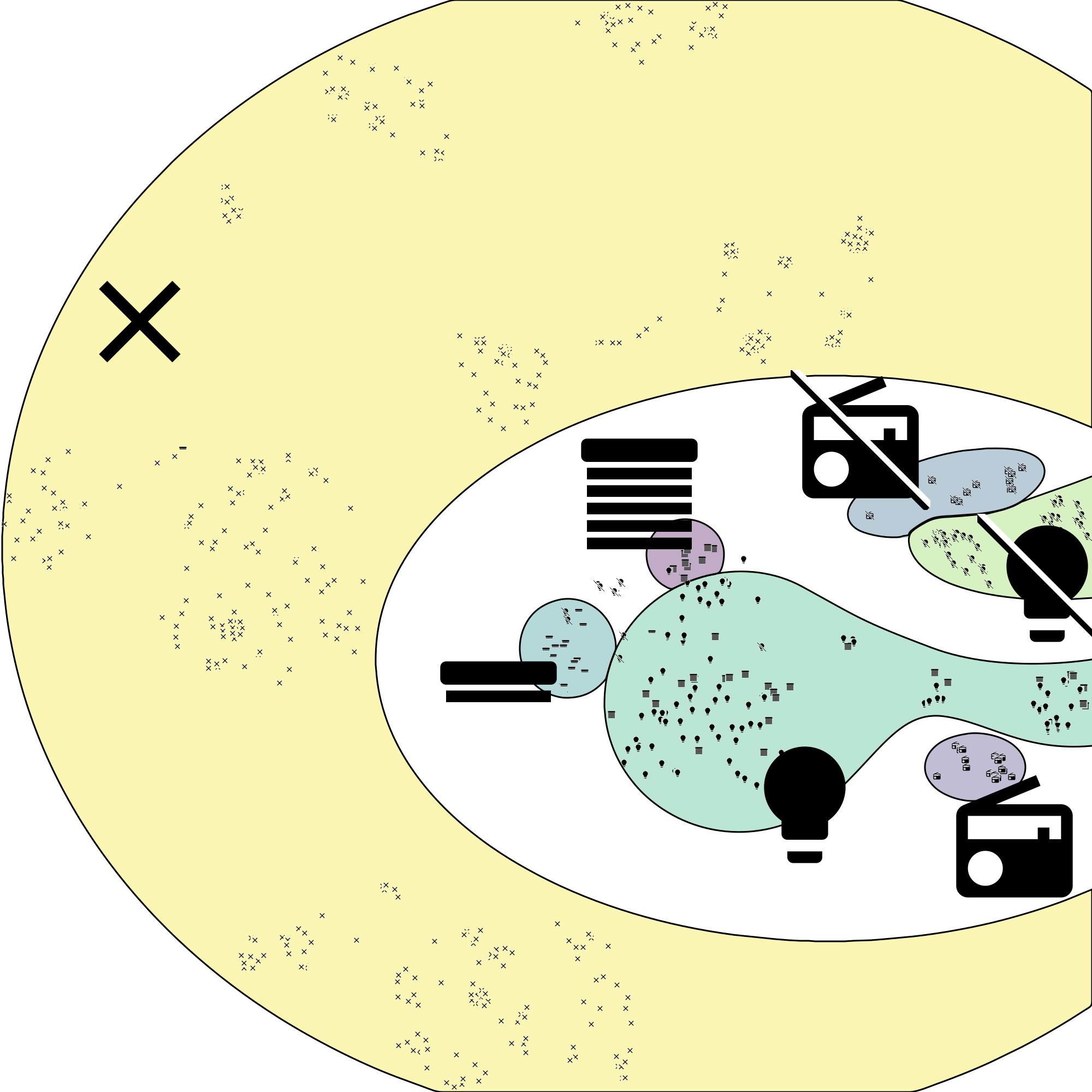}}
    \caption{Post-convolution: uttered command}
    \label{fig:complementary:tsne:raw:conv:command}
  \end{subfigure}
  \bigbreak{}

  \begin{subfigure}[b]{0.49\columnwidth}
    \centering
    \fbox{\includegraphics[width=0.99\columnwidth]{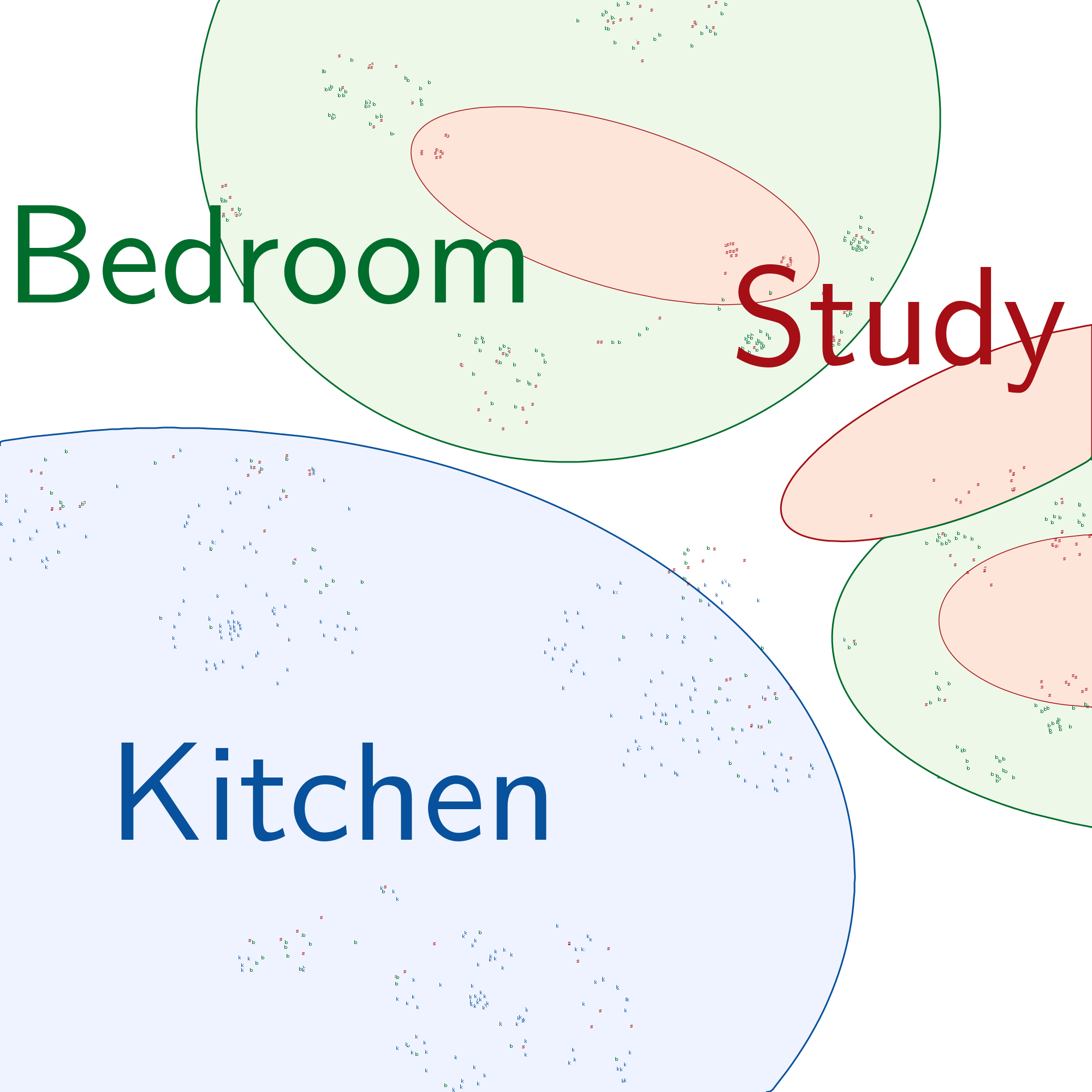}}
    \caption{Post-convolution: user location}
    \label{fig:complementary:tsne:raw:conv:location}
  \end{subfigure}
  \hfill
  \begin{subfigure}[b]{0.49\columnwidth}
    \centering
    \fbox{\includegraphics[width=0.99\columnwidth]{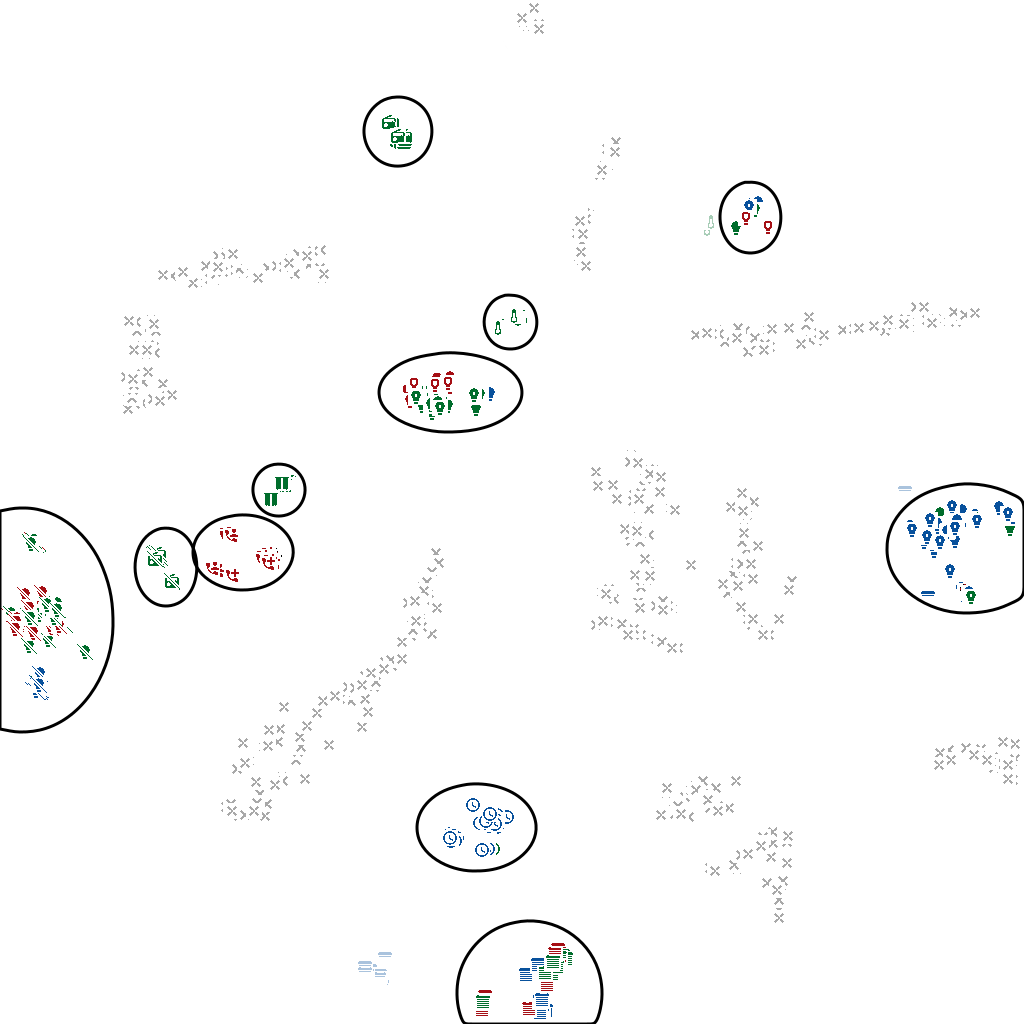}}
    \caption{Pre-classification: expected action}
    \label{fig:complementary:tsne:raw:out}
  \end{subfigure}\par

  \caption[t-SNE projections]{t-SNE projections}
  \label{fig:complementary:tsne:raw}
\end{figure*}
\typeout{^^Jt-SNE... DONE^^J}

Results presented in \autoref{fig:complementary:tsne:raw}(\subref{fig:complementary:tsne:raw:conv:location}, \subref{fig:complementary:tsne:raw:conv:command}), show that some information are identified and extracted by the convolutional part of the network.
For example, coherent blobs can be spotted for the uttered command and the user location.
By contrast, no group can be made for the activity (\autoref{fig:complementary:tsne:raw:conv:activity}), which does not seem to be a relevant feature to extract from the image for the DQN.
These results confirm the usual use of the user location as a contextual information.
However, the user's activity is not exploited by the network. It might be possible that a lower level annotations, such as agitation level, could be extracted.

\typeout{^^Jt-SNE out raw...^^J}

\typeout{^^Jt-SNE out raw... DONE^^J}

The last projection, presented in \autoref{fig:complementary:tsne:raw:out}, provides information about the way the neural network transforms the space to make each class linearly separable.
While there is a majority of \enquote{Do Nothing} action, we can identify some groups of common action type (icon type) and action place (icon color), proving that given states contain enough information to classify correctly most of them.
\clearpage{}
  \clearpage{}\section{Discussion}
\label{sec:discussion}

The results of the experiments of a reinforcement learning method with a deep model for automatic adaptive decision-making in smart-homes show promises.
In particular, the experiments show that:
\begin{itemize}
  \item the system is able to use trivial user's interactions (rewards) to adapt a decision-making process;
  \item graphical representation of multi-modal heterogeneous data can be interpreted to make decisions;
  \item the system can adapt its behavior to a user and to faulty or uninstalled sensors;
  \item the deep model can learn relevant contextual information;
  \item learning can be done on-line, using a limited history of the past and no information about the future;
  \item the method can scale to big input space (raw sensor data).
\end{itemize}

All the work has been conceived to be reproducible and all the written source code as well as the graphic components are available on-line\footnote{\myProjectWeb{}}, as a set of Lua source code and SVG files.
All these sources have been put under version control system (Git).
This allows the reproduction of the experiments, freezing and tagging the source for each experiment.
The source code follows a modular approach for improving re-usability of each component.
Regarding the data used for the experiments, it can be retrieved separately\footnote{\url{http://sweet-home-data.imag.fr/}}.

If \ARCADES{} exhibits interesting results, it opens new research questions and possible improvements.

At the RL level, the reward function plays a very important role in driving the learning in an acceptable solution.
In our previous study \parencite{BrenonEtAl2016} different reward functions were tried, taking account of the semantic distance between two actions (e.g. lighting a lamp instead of another) or the real distance between two action locations (lighting in the kitchen rather than the study).
Nevertheless, these more sophisticated functions did not bring a significant improvement in the agent behavior that is why we decided to stick to a more simple one.
Though this function led to better learning, it is worth investigating other reward function that would be closer to human evaluation (e.g. a confusion between a lamp in a room should be less problematic than a confusion of room).
Similarly, in a real interactive setup, microphones can be used to acquire a more nuanced evaluation from the user, for example using a set of keywords like \enquote{No!}, \enquote{Very good.}, \enquote{Keep going.}, etc.
Another research direction would be to take into account implicit rewards from the user as done by \textcite{KaramiEtAl2016}.

At a deep learning level, it is well-known that these models require an unreasonable quantity of data as well as a long time before reaching the convergence point.
These are important factors for acceptability of the system by naive user and should be reduced as much as possible.
If the pre-training phase is a first solution, others should be studied, such as the use of an expert system during the first interactions, waiting for \ARCADES{} to learn an appropriate model.
Recent works propose different methods to leverage these parameters.
For instance, \textcite{PritzelEtAl2017} make an intensive use of few relevant samples to reduce the number of the overall required interactions.
In parallel, \textcite{UsunierEtAl2016} propose a different strategy than the $\epsilon$-greedy one, which speeds up the exploration of the different actions.
Knowledge transfer could be another alternative which need more experiments to be validated.

Finally, paradigm shifts can also be interesting.
The use of MDP as the underlying model of RL tasks is related to the computation complexity induced by more generic models like the POMDP \parencite{Cassandra2016}.
However, these models are more suitable to handle the unknown and variable behavior of the user.
The DQN approach being able to scale to big state space, it can be interesting to try to use more generic models like POMDP or MOMDP \parencite{Araya-LopezEtAl2010} as the underlying one of our RL task.
More complex neural network architectures can also be implemented such as Recurrent Neural Networks (RNN, \cite{LiptonEtAl2015}), allowing the system to better handle time dependency as well as implementing some attention mechanisms \parencite{XuEtAl2015}.

Despite this need for improvements, this study show that deep reinforcement learning is feasible and shows promises for a life long adaptive and robust context aware decision making.
\clearpage{}
  
  \clearpage{}\bibliographystyle{elsarticle-harv.bst}
\bibliography{./assets/bibliography}\clearpage{}
  \clearpage{}\appendix{}

\section{Hyper-parameters optimization}
\label{sec:appendix:hyperoptim}

Through the whole available hyper-parameters of the system, we choose to make an exhaustive search for two of them.
The \targetq{} parameter defines the number of interactions during which a frozen network is used to compute the target $Q$-values before being updated with the current $Q$-network.
The \minibatchsize{} parameter defines the number of samples drawn to build a mini-batch used for training the network.

\textcite{MnihEtAl2015} state that these two parameters are very important to provide a good stability of the performances.
Nevertheless, as shown on the bottom left plot of the \autoref{fig:hyperoptim:1}, the \minibatchsize{} parameter heavily impact the execution time.
In spite of improving the average reward per episode and F1-score, using a high \minibatchsize{} make the system slower to process the same number of interactions.
The bottom right plot provides an intuitive view of a compromise, relating the F1-score to the execution time.
This lead us to choose a \targetq{} of $4096$ with a \minibatchsize{} of $16$.

However, another hyper-parameter, called \updatefreq{}, allows to reduce the execution time.
The learning step of the agent is indeed made each \updatefreq{} interactions.
Thus, increasing this number will execute the time-spending step less often, and yet need bigger mini-batches to reach the same level of performance.
The \autoref{fig:hyperoptim:2} presents the same plots but with different hyper-parameters.
As expected, a low \updatefreq{} increases the execution time.
The plot of the score-time ratio lead us to choose a bigger \minibatchsize{} parameter of $32$, increasing the \updatefreq{} parameter to $12$.

The old and new values of hyper-parameters are summed up in the \autoref{tab:hyperoptim:sumup}.

\begin{table}[!ht]
\centering
\caption{Evolution of the hyper-parameters}
\label{tab:hyperoptim:sumup}
\begin{tabular}{rccc} \toprule
 & Initial value & Intermediate value & Final value \\
\midrule
\verb|target_q| & $1024$ & $4096$ & $4096$ \\
\verb|minibatch_size| & $64$ & $16$ & $32$ \\
\verb|update_freq| & $3$ & $3$ & $12$ \\
\bottomrule
\end{tabular}
\end{table}

\setlength{\belowcaptionskip}{0pt}
\setlength{\abovecaptionskip}{0pt}
\begin{figure*}[bh]
  \begin{minipage}[c]{0.45\textwidth}
    \includeplot[0.99\columnwidth]{./assets/gnuplot/hyperoptim}{1}
    \caption{Metrics related to different choices of hyper-parameters}
    \label{fig:hyperoptim:1}
  \end{minipage}
  \hfill
  \rule[-50pt]{0.1pt}{100pt}
  \hfill
  \begin{minipage}[c]{0.45\textwidth}
    \includeplot[0.99\columnwidth]{./assets/gnuplot/hyperoptim}{2}
    \caption{Metrics related to different choices of hyper-parameters}
    \label{fig:hyperoptim:2}
  \end{minipage}
\end{figure*}
\clearpage{}
\end{document}